\documentclass[10pt,twocolumn,letterpaper]{article}

\PassOptionsToPackage{table}{xcolor}
\usepackage{wacv}              

\usepackage{multirow}
\usepackage{makecell}
\usepackage{tcolorbox}
\usepackage{pifont}

\definecolor{wacvblue}{rgb}{0.21,0.49,0.74}
\usepackage[pagebackref,breaklinks,colorlinks,allcolors=wacvblue]{hyperref}

\def\wacvPaperID{*****}
\def\confName{WACV}
\def\confYear{2027}

\title{ObsDriveBench: Benchmarking Multimodal Understanding under Adverse Weather with Observability Awareness}

\author{
Qiao Yan\textsuperscript{1} \quad
Yihan Wang\textsuperscript{1} \quad
Zhenghao Xing\textsuperscript{1} \quad
Jiaqi Xu\textsuperscript{1,*} \quad
Pheng-Ann Heng\textsuperscript{1,2}
\\[3pt]
\textsuperscript{1}Department of Computer Science and Engineering, 
The Chinese University of Hong Kong
\\
\textsuperscript{2}Institute of Medical Intelligence and XR,
The Chinese University of Hong Kong
\\
\textsuperscript{*}Corresponding author
}

\begin{document}
\maketitle

\begin{abstract}
Autonomous driving under adverse weather remains a critical challenge, yet existing vision-language benchmarks mainly evaluate under standard conditions, synthetic corruptions, or single modality. As a result, it remains unclear how vision-language models behave under real-world adverse weather with multi-modal inputs. 
We argue that a key difficulty lies in degraded environmental observability: under fog, rain, snow, and low illumination, multi-modal observations become unreliable and cross-modally inconsistent, posing challenges to scene understanding, and subsequent decision-making. 
To study this, we introduce \textbf{ObsDriveBench}, a real-world multi-modal benchmark for adverse-weather autonomous driving. Our benchmark is designed with three capability dimensions: \textbf{observability awareness}, \textbf{spatial reliability}, and \textbf{risk-aware decision-making}, enabling fine-grained diagnosis of model behavior under degraded observations. 
We construct the benchmark through observability meta-annotation, scene description, and capability oriented multiple-choice tasks over synchronized camera, LiDAR, and radar inputs, forming a benchmark with over 14k training and 13k test questions. 
Experiments reveal consistent performance degradation of existing vision-language models. We further introduce \textbf{ObsDrive} model with normal-weather supervised fine-tuning and adverse-weather reinforcement learning, improving robustness across all three capabilities. The dataset and evaluation code will be released at \href{https://github.com/russellyq/ObsDriveBench}{\texttt{ObsDriveBench}}.
\end{abstract}



\section{Introduction}
\label{sec_intro}


Autonomous driving under adverse weather has been extensively studied, primarily focusing on module-level robustness, such as detection, segmentation, and multi-modal sensor fusion. 
Recent driving vision-language benchmarks, such as DriveLM \cite{sima2024drivelm}, DriveLMM-o1 \cite{ishaq2025drivelmm}, NuScenes-SpatialQA \cite{tian2025nuscenes}, STSBench \cite{fruhwirth2025stsbench}, OmniDrive \cite{wang2025omnidrive}, and FutureVQA \cite{chang2026probing}, have begun to evaluate higher-level capabilities, including spatial understanding, interaction prediction, and decision-making. However, these benchmarks are largely constructed under standard driving conditions and do not explicitly assess model behavior under real-world adverse weather. Recently, some works start to explore robustness under degraded conditions: DriveBench \cite{xie2025vlms} introduces synthetic image corruptions, showing that VLM-based driving systems exhibit noticeable performance drops under visual noise. RoboDriveVLM \cite{liao2025robodrivevlm} further extends this setting by incorporating multi-modal sensor corruptions, including lidar and radar degradation. However, it focuses primarily on end-to-end trajectory prediction, without explicitly analyzing intermediate understanding and reasoning capabilities. Although multi-modal sensing has been widely adopted to improve perception robustness under adverse weather, its role in supporting high-level understanding and decision-making remains underexplored.

As a result, a key question remains: \emph{how do vision-language models behave under real-world adverse weather, and how can we systematically evaluate their limitations?} We argue that the main challenge lies in \textbf{degraded environmental observability}. Under adverse weather conditions such as fog, rain, snow, the environment becomes inherently \emph{less observable}. Camera inputs suffer from reduced visibility, LiDAR returns become sparse and unstable due to scattering, while radar, although more robust, provides only coarse geometric cues. Consequently, multi-modal observations become both degraded and cross-modally inconsistent, posing challenges to reliable scene understanding. Such degradation affects multiple aspects of the driving pipeline. Under limited and inconsistent observations, models may struggle to maintain reliable spatial understanding, including object grounding, relative positioning. These limitations can further impact downstream tasks such as interaction prediction and decision-making, especially in safety-critical scenarios. Therefore, the key issue is whether models maintain consistent and reliable understanding under degraded observations.

\begin{figure}[t!]
    \centering
    \includegraphics[width=0.99\linewidth]{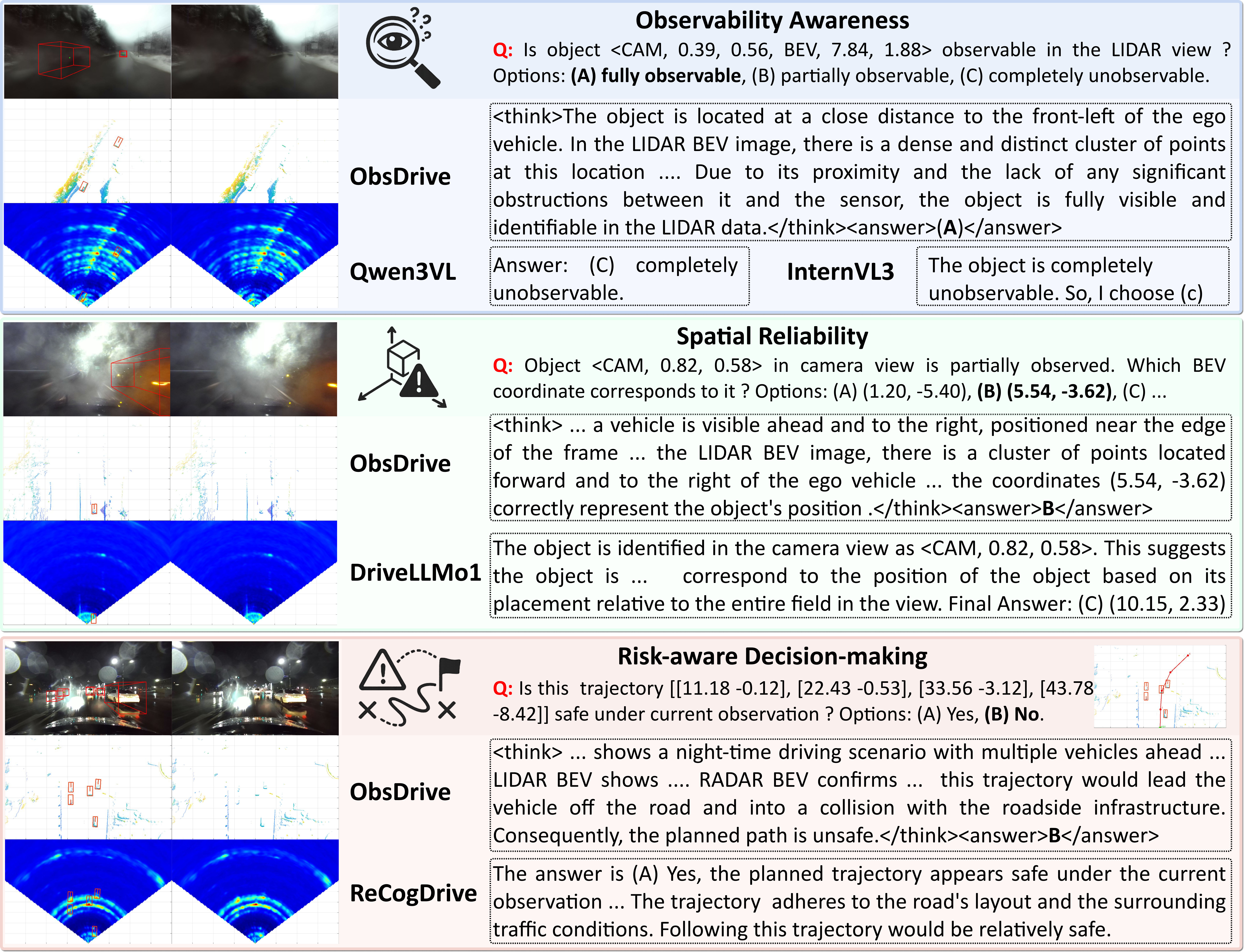}
      \caption{
    Examples from ObsDriveBench. Compared with existing models, ObsDrive demonstrates more consistent understanding across modalities and tasks.
    }
    \label{fig:intro}
\end{figure}

To systematically study this problem, we introduce a real-world multi-modal benchmark, namely \textbf{ObsDriveBench}, for evaluating autonomous driving under adverse weather. ObsDriveBench provides a structured evaluation of model behavior along three complementary capability dimensions: \textbf{observability awareness}, which measures the ability to recognize degraded visibility, partial observations, and cross-modal inconsistencies and verification; \textbf{spatial reliability}, which evaluates whether models maintain consistent and plausible spatial understanding under degraded observations; and \textbf{risk-aware decision-making}, which assesses whether models adapt their behavior and make safe decisions under uncertainty. 
These dimensions provide a structured framework to analyze how limitations in perception and understanding manifest across the driving pipeline.
ObsDriveBench is built upon synchronized camera, LiDAR, and radar data, and integrates \emph{observability-aware meta-annotations} with \emph{capability-oriented task design}. Specifically, we generate structured multiple-choice questions grounded in multi-modal inputs and object-level annotations, covering modality-specific observability estimation, cross-modal observability comparison and verification, 2D–3D grounding, spatial relation consistency, interaction reasoning, as well as risk identification, fallback action, and trajectory safety evaluation. This design enables fine-grained and interpretable diagnosis of model behavior under adverse weather, providing insights into where and how models fail. We summarize our main contributions as follows:

\begin{itemize}

    \item We introduce \textbf{ObsDriveBench}, an observability-driven benchmark for multimodal driving under real adverse weather. It evaluates the failure-propagation chain from observability awareness to spatial reliability and risk-aware decision-making, enabling diagnosis that cannot be obtained from final trajectory metrics alone. Its training split further provides observability annotations, scene descriptions, and capability-oriented supervision for developing future driving VLMs.
    

    \item We conduct extensive evaluations of state-of-the-art models, revealing several systematic limitations under adverse weather. In particular, models exhibit limited ability to accurately estimate observability across modalities, struggle to maintain consistent spatial understanding under degraded observations, and fail to reliably identify risky regions and evaluate trajectory safety. Notably, while many models tend to adopt conservative fallback actions, they often lack precise understanding of where risks originate and which trajectories are truly safe.

    \item We develop \textbf{ObsDrive} model with a two-stage training strategy, including supervised fine-tuning on normal-weather data and reinforcement learning on adverse-weather scenarios. This design improves robustness across all three capability dimensions, demonstrating the effectiveness in multimodal driving understanding.
\end{itemize}

\section{Related Works}
\label{sec_related_works}

\subsection{Adverse Weather Perception}

Perception under adverse weather has been extensively studied in autonomous driving, focusing on improving robustness of sensing modalities such as cameras, LiDAR, and radar. Adverse conditions, including fog, rain, snow, and low illumination, significantly degrade sensor observations. Image-based perception suffers from reduced contrast and noise \cite{liu2026robust}, LiDAR becomes sparse and unstable due to scattering \cite{dreissig2023survey}, while radar, although more robust, provides only coarse geometric information \cite{han20234d}. To address these challenges, prior works have explored image enhancement, point cloud denoising, and multi-modal fusion, leveraging complementary properties of different sensors \cite{zhang2023perception}. Several datasets, such as RADIATE \cite{sheeny2021radiate} and K-Radar \cite{paek2022k}, further support research under adverse weather conditions.  However, most existing efforts focus on \emph{low-level perception tasks}, such as detection and segmentation, treating adverse weather as a sensing problem. The impact of degraded and cross-modally inconsistent observations on downstream reasoning and decision-making remains largely underexplored.

\begin{figure*}[ht!]
    \centering
    \includegraphics[width=0.99\linewidth]{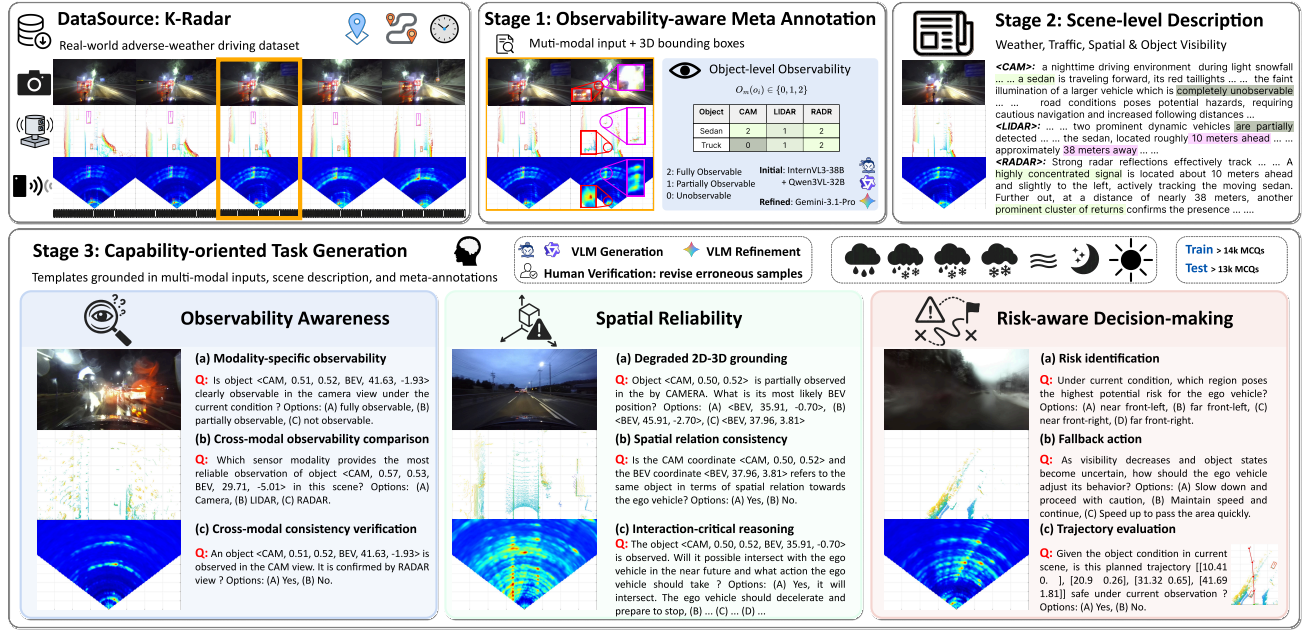}
    \caption{Pipeline of benchmark construction. Sampled from the real-world adverse-weather driving dataset, we first perform \textbf{observability-aware meta-annotation} on synchronized camera, lidar, and radar inputs with 3D bounding boxes, producing object-level observability labels across modalities. We then generate \textbf{scene-level descriptions} that summarize weather, traffic, spatial relations, and visibility cues under degraded observations. Based on the multi-modal inputs, observability annotations, and scene descriptions, we construct \textbf{capability-oriented MCQ tasks} covering three dimensions: \textbf{observability awareness}, \textbf{spatial reliability}, and \textbf{risk-aware decision-making}.}
    \label{fig:data_gen}
\end{figure*}

\subsection{Vision-Language Models for Autonomous Driving}

Recent vision-language models (VLMs) and vision-language-action (VLA) frameworks introduce a new paradigm for autonomous driving, enabling unified reasoning across perception, prediction, planning, and behavior. Benchmarks such as DriveLM \cite{sima2024drivelm}, DriveLMM-o1 \cite{ishaq2025drivelmm}, NuScenes-QA \cite{qian2024nuscenes}, NuScenes-SpatialQA \cite{tian2025nuscenes}, STSBench \cite{fruhwirth2025stsbench}, OmniDrive \cite{wang2025omnidrive}, and FutureVQA \cite{chang2026probing} evaluate high-level reasoning capabilities, including spatial understanding, spatio-temporal interaction, and counterfactual reasoning. VLA frameworks \cite{li2025spacedrive,li2026recogdrive,zhou2025opendrivevla,zhang2025omnidrive,sun2025minddrive,wang2025alpamayo} further extend this paradigm to end-to-end trajectory prediction. Recent works begin to study robustness under degraded conditions. DriveBench \cite{xie2025vlms} and RoboDriveChallenge \cite{kong2024robodrive} introduce synthetic image corruptions, revealing significant performance degradation, while RoboDriveVLM \cite{liao2025robodrivevlm} incorporates multi-modal sensor corruptions. However, these approaches either focus on image-level perturbations or end-to-end trajectory prediction, without explicitly evaluating reliability across the full driving pipeline. Moreover, existing VLM/VLA methods largely treat multi-modal inputs as auxiliary signals for perception, rather than explicitly modeling how different modalities contribute to reasoning under degraded and inconsistent observations. As a result, current benchmarks do not capture how observability degradation propagates through spatial reasoning and affects safety-critical decision-making. In contrast, our work explicitly models this failure mode from observability degradation to spatial unreliability and unsafe behavior, providing a new benchmark for evaluating understanding and reasoning reliability under adverse weather.
ObsDriveBench complements rather than replaces continuous trajectory-evaluation benchmarks. End-to-end metrics reveal whether a final trajectory is accurate or safe, but do not localize whether failure originates from unreliable observations, cross-modal inconsistency, spatial grounding, or risk assessment. Our capability-oriented MCQs isolate these intermediate failure sources along the observability-spatial-decision chain. We therefore evaluate decision-relevant understanding and do not claim that MCQ accuracy alone constitutes closed-loop planning performance.


\section{Benchmark Construction}
\label{sec_benchmark}

\subsection{Overview}

Our benchmark is designed to evaluate autonomous driving systems under adverse weather through a structured evaluation framework that characterizes model behavior under degraded and uncertain observations. We organize the benchmark along three complementary capability dimensions:
\textbf{Observability Awareness}: the ability to explicitly estimate modality-dependent observability and identify cross-modal inconsistencies and verification under degraded observations;
\textbf{Spatial Reliability}: the ability to maintain geometrically consistent and physically plausible spatial understanding under partial and uncertain observations, including cross-view grounding, spatial relations, and interaction reasoning;
\textbf{Risk-aware Decision-making}: the ability to adapt decision-making by accounting for uncertainty in observation and spatial understanding, and to select safety-aligned actions, including identifying risk-critical regions, adopting conservative fallback strategies, and evaluating trajectory safety.
To evaluate these capabilities, we construct a multimodal benchmark, namely \textbf{ObsDriveBench} that integrates real-world adverse-weather driving data, object-level observability annotations, and capability-oriented task generation. Our benchmark supports fine-grained and interpretable analysis of model behavior under degraded observations across different aspects of multimodal understanding.
Details on dataset construction, weather and daytime/nighttime distributions, and observability meta-annotation are provided in the supplementary material.

\subsection{Data Preparation}
\label{sec_data_prepration}
We build our benchmark on K-Radar~\cite{paek2022k}, a large-scale real-world autonomous-driving dataset containing diverse adverse-weather scenarios and synchronized multimodal observations, including camera, LiDAR, and radar. To provide these modalities to VLMs through a unified visual interface, we retain the original camera images and render the LiDAR and radar measurements as bird's-eye-view (BEV) images.
Each sample is constructed as a temporal sequence of 21 frames, covering a 1-second past context and a 1-second future horizon, enabling both current scene understanding and future-aware reasoning during the data construction process. \noindent\textbf{Train-test split.} We inherit the official K-Radar split used for object detection, rather than randomly splitting frames or generated questions. Each sample and every MCQ derived from the inherit split of source sequence. 

\noindent\textbf{Stage 1: Observability-aware Meta-annotation.}
We explicitly model cross-modal object-level observability instead of assuming all objects are equally observable. Given an object $o_i$ and modality $m \in \{\text{cam}, \text{lidar}, \text{radar}\}$, we define:
$O_m(o_i) \in \{0,1,2\}$
where $2$, $1$, and $0$ denote fully observable, partially observable, and unobservable, respectively. 
To obtain these annotations, we leverage strong vision-language models (InternVL3-38B~\cite{zhu2025internvl3} and Qwen3VL-32B-Instruct~\cite{Qwen3-VL}) to analyze multimodal inputs with 3D bounding boxes and generate initial observability labels. These annotations are further refined using Gemini-3.1-Pro~\cite{gemini31pro} to improve consistency and reliability.
\textbf{Stage 2: Scene-level Descriptions.}
Based on multimodal inputs and observability annotations, we generate structured scene descriptions that summarize object states, spatial relations, and interaction cues under degraded observations. This step bridges low-level perception and high-level understanding and reasoning.
\textbf{Stage 3: Capability-oriented Task Generation.}
We construct multiple choice questions (MCQs) from parameterized templates grounded in both scene descriptions and meta-annotations, together with camera views and BEV of LiDAR and radar. These questions are explicitly designed to target three capability dimensions: observability awareness, spatial reliability, and risk-aware decision-making. All questions are initially generated using InternVL3-38B and Qwen3VL-32B-Instruct, followed by refinement with Gemini-3.1-Pro to improve clarity and correctness. Finally, we perform human verification to correct erroneous samples, ensuring high-quality evaluation data.
\noindent\textbf{Human verification.}
All generated MCQs undergo human verification by PhD-level annotators in computer science and engineering. Annotators are provided with the synchronized camera, LiDAR, and radar views, original K-Radar 3D object annotations, ground-truth ego trajectories, and generated MCQs. They verify that each question refers to valid scene content, has a unique answer consistent with the sensor observations and ground truth, and contains no spatial or temporal inconsistency. Erroneous samples are corrected. The final verification pass rates are 99.96\% for observability awareness, 97.20\% for spatial reliability, and 99.96\% for risk-aware decision-making.

\subsection{Task Generation}
Building upon the observability-aware meta-annotations and scene-level descriptions, we construct a structured set of tasks that operationalize three complementary capability dimensions: \textbf{observability awareness}, \textbf{spatial reliability}, and \textbf{risk-aware decision making}. We organize tasks into capability-oriented families, each designed to probe a specific aspect of model behavior under degraded and cross-modally inconsistent observations. Each task is instantiated from parameterized templates grounded in multimodal inputs and object-level meta-annotations, enabling systematic diagnosis of model behavior.

\noindent\textbf{(1) Observability Awareness}
These tasks require models to explicitly estimate modality-dependent observability and to understand cross-modal inconsistency. We construct three categories: \textit{(a) Modality-specific observability}, where the model predicts the observability level of an object under a given modality (camera, LiDAR, or radar), capturing degradation effects such as occlusion, sparsity, and noise. \textit{(b) Cross-modal observability comparison}, where the model identifies which modality provides the most reliable observation, reflecting modality-dependent robustness under adverse weather. \textit{(c) Cross-modal observability verification}, where the model evaluates whether observations from different modalities are consistent, explicitly modeling disagreement and uncertainty across sensors. These tasks explicitly require models to understand about \emph{what can be reliably observed} before making further inferences.

\noindent\textbf{(2) Spatial Reliability under Uncertainty.} Given degraded and uncertain observations, these tasks evaluate whether models can maintain consistency in spatial understanding. We construct three types of tasks: \textit{(a) Degraded 2D–3D grounding}, where models associate objects between camera views and BEV representations, testing 2D-to-3D grounding capability. \textit{(b) Spatial relation consistency}, where models determine whether spatial relations inferred from different representations (e.g., camera vs BEV) are geometrically consistent, emphasizing cross-view spatial coherence. \textit{(c) Interaction-critical reasoning}, where potential interactions are modeled with ego vehicle under uncertain observability, requiring integration of spatial and motion cues. These tasks emphasize that under adverse weather, the key challenge is whether spatial understanding remains \emph{reliable under uncertainty}.

\noindent\textbf{(3) Risk-aware Decision-making.}
At the decision level, we evaluate whether models can adapt behavior when perception and spatial understanding become unreliable. We design three categories: \textit{(a) Risk identification}, where models identify high-risk objects or regions under limited observability, requiring implicit uncertainty-aware risk estimation. \textit{(b) Fallback action}, where models behave with conservative actions when visibility decreases or object states become uncertain, reflecting safe behavior. \textit{(c) Trajectory evaluation}, where models assess the safety of candidate trajectories under uncertain observations, requiring reasoning about potential future outcomes. Noticeably, trajectory evaluation includes both \emph{feasible} and \emph{counterfactual} trajectories, requiring models to distinguish between safety-aligned and risk-inducing behaviors. (detailed in the supplementary material). These tasks explicitly focus \emph{decision-making under uncertainty}, explicitly evaluating whether models adopt safety-aligned behavior when observations are unreliable.

\begin{figure}[t!]
    \centering
    \includegraphics[width=0.99\linewidth]{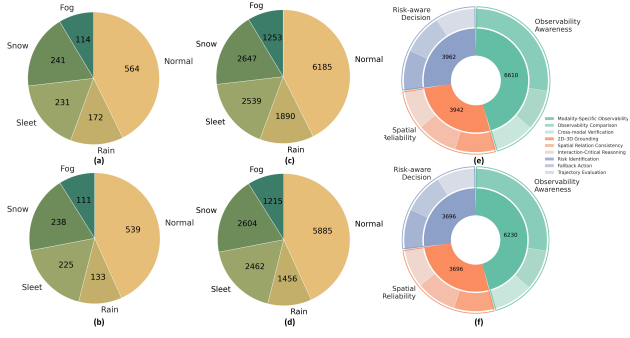}
    \caption{Data distribution of ObsDriveBench. Frame-level weather distribution of the (a) train and (b) test splits. Sample-level (MCQ) weather distribution of the (c) train and test (d) splits. Hierarchical task composition of the (e) train and (f) test splits, showing the three major capability groups, observability awareness, spatial reliability, and risk-aware decision-making, together with their sub-task breakdowns.}
    \label{fig:data_distribution}
\end{figure}

\noindent\textbf{Discussion.}
The proposed task categories provide a structured perspective to analyze model behavior under adverse weather. By organizing evaluation across observability awareness, spatial reliability, and risk-aware decision-making, our benchmark enables fine-grained diagnosis of model limitations across different aspects of the driving pipeline. Full prompt templates and more examples are provided in the supplementary material.

\begin{table*}[t!]
\caption{Performance across three task capability dimensions. AVG denotes the average score within each group. The best is \textbf{bold} and the second-best is \underline{underlined}.}
\label{tab:main_results}
\resizebox{\textwidth}{!}{%
\begin{tabular}{lc|cccc|cccc|cccc}
\toprule
\multirow{2}{*}{\textbf{Model}} & \multirow{2}{*}{\textbf{Venue}} 
& \multicolumn{4}{c|}{\textbf{Observability Awareness}} 
& \multicolumn{4}{c|}{\textbf{Spatial Reliability}} 
& \multicolumn{4}{c}{\textbf{Risk-aware Decision-making}} \\

& 
& \makecell{Modal \\ Obs.} 
& \makecell{Cross-mod. \\ Comp.} 
& \makecell{Cross-mod. \\ Verif.} 
& \textbf{AVG} 

& \makecell{2D-3D \\ Ground.} 
& \makecell{Spatial \\ Relation} 
& \makecell{Interaction \\ Reasoning} 
& \textbf{AVG} 

& \makecell{Risk \\ Ident.} 
& \makecell{Fallback \\ Action} 
& \makecell{Traj. \\ Eval.} 
& \textbf{AVG} \\
\midrule \midrule

\multicolumn{14}{c}{\textit{Close-source Proprietary Models}} \\ \midrule\midrule

GPT-5 & 2025-11 & 43.55 & 19.42 & 59.39 & 41.89 & 62.01 & 58.43 & 62.74 & 61.04 & 53.30 & 92.22 & 54.48 & 66.80 \\
Gemini-3.1-Pro & 2026-02 & \underline{54.49} & 27.69 & \textbf{82.42} & 54.72 & \textbf{89.08} & \textbf{89.81} & \textbf{85.15} & \textbf{88.04} & \textbf{68.33} & \textbf{99.76} & 72.31 & 80.22 \\
\midrule\midrule
\multicolumn{14}{c}{\textit{Open-source General Models}} \\ \midrule\midrule

InternVL25-7B & 2024-12 & 35.69 & 29.86 & 54.41 & 38.27 & 29.80 & 53.45 & 59.34 & 47.40 & 50.80 & 95.10 & 50.25 & 65.56 \\
InternVL3-8B & 2025-04 & 43.10 & 30.26 & 64.45 & 44.80 & 38.80 & 50.40 & 53.94 & 47.65 & 52.81 & 91.25 & 49.67 & 64.75 \\
InternVL3-38B & 2025-04 & 38.68 & 30.34 & 57.70 & 40.82 & 29.64 & 50.80 & 62.16 & 47.38 & 42.93 & 80.74 & 49.75 & 57.90 \\

Qwen2.5VL-7B & 2025-02 & 41.31 & 34.27 & 60.51 & 43.74 & 44.98 & 34.91 & 67.39 & 48.89 & 55.14 & 94.70 & 52.16 & 67.51 \\
Qwen3VL-8B & 2025-12 & 13.51 & 40.05 & 28.41 & 21.80  & 46.18 & 30.50  & 63.65 & 46.59 & 55.14 & 99.52 & 60.28 & 71.78 \\
Qwen3VL-32B & 2025-12 & 34.16 & \underline{44.06} & 60.67 & 41.44 & 62.17 &  60.91 & 78.59  & 67.10 & 60.13 & \textbf{99.76} & 79.19 & 79.71 \\
 \midrule \midrule

\multicolumn{14}{c}{\textit{Open-source Reasoning Models}} \\ \midrule\midrule
R1-VL-7B & ICCV2025 & 27.88 & 24.80 & 26.40 & 26.97 & 26.75 & 29.13 & 55.35 & 36.88 & 53.78 & 96.90 & 55.47 & 68.59 \\
Perception-R1-7B & ICLR2026 & 46.42 & 33.39 & 74.48 & 49.42 & 44.26 & 59.95 & 65.06 & 56.33 & 50.16 & 98.64 & 53.07 & 67.45 \\
\midrule \midrule

\multicolumn{14}{c}{\textit{Open-source Expert Models}} \\ \midrule\midrule
Dolphins & ECCV2024 & 40.07 &30.82& 29.53 & 36.12 & 24.98 & 29.37 & 33.94 & 29.38 & 19.05 & 30.82 & 54.73 & 34.66 \\
DriveLMM-o1 & IROS2025 & 35.02 & 33.23 & 56.02 & 38.86 & 30.76 & 49.52 & 63.40 & 47.73 & 53.94 & 93.66 & 49.25 & 65.80 \\
Alpamayo-1.5 & arXiv2025 & 12.20 & 8.19 & 26.48 & 14.25 & 7.63 & 28.33 & 62.82 & 32.60 & 56.75 & 99.60 & 42.45 & 66.53 \\
ReasonDrive & CVPRW2025 & 39.38 & 14.61 & 73.92 & 41.33 & 85.62 & 78.81 & 65.56 & 76.79 & 48.55 & 49.52 & 41.38 & 46.54 \\
ReCogDrive & ICLR2026 & 35.26 & 32.58 & 52.73 & 38.22 & 39.36 & 51.44 & 69.29 & 53.19 & 52.49 & 96.31 & 66.09 & 71.70 \\

\midrule\midrule
\rowcolor[HTML]{EFEFEF}ObsDrive-SFT & - & 53.53 & 41.09 & 73.27 & \underline{54.99} & 80.40 & 77.69 & 81.33 & 79.79 & 57.88 & 98.96 & \underline{98.09} & \underline{84.85} \\
\rowcolor[HTML]{EFEFEF}ObsDrive-RL & - & \textbf{66.32} & \textbf{46.63} & \underline{80.98} & \textbf{65.31} & \underline{85.78} & \underline{82.34} & \underline{83.40} & \underline{83.85} & 61.33 & \underline{99.36} & \textbf{99.00} & \textbf{86.44} \\

\bottomrule
\end{tabular}%
}
\end{table*}

\newif\ifcomment
\commentfalse  

\ifcomment

\subsection{Evaluation}
\label{sec_evaluation}

In addition to standard accuracy on MCQ tasks for choosing the correct option, we introduce observability-aware evaluation metrics to explicitly measure how models reason under degraded and inconsistent observations. \textbf{Observability Confusion Distance (OCD):}
We measure prediction error as the absolute distance between predicted and ground-truth observability:
\begin{equation}
\text{OCD} = \frac{1}{3N} \sum_{i=1}^{N} \sum_{m \in \{\text{cam}, \text{lidar}, \text{radar}\}} \left| O_m(o_i) - \hat{O}_m(o_i) \right|,
\end{equation}
where $N$ is the number of ground-truth objects, and the factor $3$ corresponds to the three sensing modalities. OCD measures the degree of observability misestimation by treating observability as an ordinal variable. Unlike standard accuracy, it captures not only whether the prediction is incorrect, but also how severely it deviates from the true observability level. The OCD metric ranges from $0$ to $2$, where lower values indicate better performance. Notably, it captures the \emph{severity} of of observability estimation errors. An OCD of $0$ means perfect observability estimation, while larger values reflect more severe misestimation. 

\fi

\subsection{Data Statistics}

Figure~\ref{fig:data_distribution} summarizes the data distribution of our benchmark across weather conditions and task categories. Panels (a) and (b) show the frame-level weather distribution for the training and test splits, respectively, while panels (c) and (d) present the corresponding sample-level (MCQ questions) weather distribution. Panels (e) and (f) further illustrate the hierarchical task composition of the training and test splits, including observability awareness, spatial reliability, and risk-aware decision-making together with their sub-tasks. Overall, the benchmark covers diverse adverse-weather scenarios and maintains a relatively balanced task composition across splits.
In total, our benchmark is constructed from over 1.4k training and 1.3k test frames sampled from K-Radar~\cite{paek2022k}, from which we derive more than 14k training and 13k test MCQ samples.



\section{Baseline Model}
\label{sec:baseline}

\textbf{Supervised Fine-Tuning (SFT).} To initialize the model with basic reasoning capability, we construct a SFT dataset using \textit{normal-weather scenarios}, including both daytime and nighttime conditions. These scenarios provide relatively reliable and consistent observations, allowing the model to learn stable understanding and reasoning patterns without severe observability degradation. The training data is derived from the same task formulation as our benchmark. For each sample, we generate structured supervision in the following format: $\texttt{<think>}$ $\text{reasoning process}$  $\texttt{</think>}$ $\texttt{<answer>}$ $\text{final answer}$ $\texttt{</answer>}$. 
This SFT stage serves as a cold-start initialization, enabling the model to acquire reliable reasoning patterns under \emph{fully or mostly observable conditions}, which is critical before handling more challenging degraded scenarios.

\noindent\textbf{Reinforcement Learning (RL).} To further improve robustness under degraded and uncertain observations, we perform reinforcement learning using \textit{adverse-weather scenarios}, including fog, snow, sleet, and rain, under both daytime and nighttime conditions. Compared to the SFT stage, these scenarios introduce significant observability degradation and cross-modal inconsistencies, making reasoning substantially more challenging. We adopt a standard Group Relative Policy Optimization (GRPO) \cite{shao2024deepseekmath} framework to optimize the model. Specifically, the reward is defined as $R = R_{\text{acc}} + R_{\text{format}}$
where $R_{\text{acc}}$ measures answer correctness, and $R_{\text{format}}$ enforces structured output in the required $\texttt{<think>} + \texttt{<answer>}$ format.

\section{Experiment}
\label{sec_experiment}

\subsection{Implementation Details}
\label{sec_implementation}

\noindent\textbf{Evaluated Models.} We evaluate a diverse set of multimodal large language models (MLLMs), including \textit{close-source proprietary models} (GPT-5 \cite{singh2025openai} and Gemini-3.1-Pro \cite{gemini31pro}), \textit{general-purpose models} and \textit{domain-specific expert models}. For general-purpose models, we include recent state-of-the-art MLLMs such as Qwen-VL \cite{Qwen3-VL, Qwen2.5-VL}, InternVL \cite{internvl2.5, zhu2025internvl3}, and reasoning-based models such as R1-VL \cite{zhang2025r1vllearningreasonmultimodal} and Perception-R1 \cite{xiao2026perceptionr}. For domain-specific models, we evaluate driving-oriented MLLMs, including Dolphins \cite{ma2024dolphins}, DriveLMM-o1 \cite{ishaq2025drivelmm}, Alpamayo-1.5 \cite{nvidia2025alpamayo}, ReasonDrive \cite{chahe2025reasondrive} and RecogDrive \cite{li2026recogdrive}, which are designed specifically for autonomous driving scenarios. All models are evaluated under their official inference settings to ensure fair comparison. 
\textbf{Baseline Model.}
We adopt Qwen2.5VL-7B-Instruct \cite{Qwen2.5-VL} as our baseline model. The model is further trained using two-stage pipeline, including supervised fine-tuning on normal-weather data and reinforcement learning on adverse-weather scenarios, namely ObsDrive-SFT and ObsDrive-RL. Training details are in the supplementary material. Answer correctness is judged using the LLM-as-Judges \cite{liu2024calibrating} approach for choosing the correct option.
\textbf{Role of ObsDrive.}
ObsDrive is a reference baseline used to test whether the diagnosed capability gaps can be reduced; it is not required for establishing the benchmark findings. All 14 off-the-shelf models in Table~\ref{tab:main_results} are evaluated without access to the ObsDriveBench training split. Their shared degradation therefore reflects limitations exposed by the benchmark rather than familiarity with K-Radar.
\textbf{Multimodal adaptation.}
For vision-centric VLMs, LiDAR and radar measurements are rendered as BEV images and supplied as additional visual inputs together with the camera image. All three inputs are processed by the model's original visual encoder without modifying its architecture. This unified representation enables modality-controlled evaluation while avoiding the introduction of model-specific fusion modules.

\subsection{Experimental Results}
\label{sec_results}

\subsubsection{Analysis.} 

Table~\ref{tab:main_results} reports model performance across three capability dimensions: observability awareness, spatial reliability, and risk-aware decision-making. Several findings can be revealed:
\textbf{(1) Observability Awareness remains the primary bottleneck.}
We first observe that observability awareness is consistently the weakest dimension across most models. In particular, modality-specific observability and cross-modal comparison exhibit relatively low performance, indicating that models struggle to accurately estimate what can be reliably observed under degraded conditions. While cross-modal verification tends to achieve higher scores, its performance is still limited. This indicates that models struggle not only with fine-grained observability estimation, but also with reliably determining cross-modal consistency. Notably, Gemini-3.1-Pro achieves strong performance on cross-modal verification, suggesting that large proprietary models have better capability in recognizing modality agreement, but still fall short in precise observability modeling. Overall, these results indicate that current MLLMs lack a fundamental understanding of observability under adverse weather.
\textbf{(2) Spatial Reliability is limited.}
Performance remains constrained in spatial reliability tasks. Models achieve suboptimal performance on 2D-3D grounding and spatial relation consistency, suggesting that they struggle to align multi-view representations. In addition, performance on interaction-critical reasoning is still limited, reflecting difficulty in integrating spatial and motion cues under uncertainty. This gap highlights that even when spatial representations are available, unreliable or inconsistent observations hinder accurate spatial understanding.
\textbf{(3) Decision-making exhibits an apparent \textit{paradox}.}
Noticeably, models achieve relatively high performance on fallback action. Most models demonstrate strong ability to select conservative actions under degraded visibility (often exceeding 90\%), indicating that they can recognize the presence of uncertainty and adopt safety-oriented behaviors. However, this capability does not translate to precise decision-making. Performance on risk identification and trajectory evaluation remains significantly lower, suggesting that models fail to accurately localize risk or evaluate future waypoints. In particular, trajectory evaluation shows a clear gap, indicating that models struggle to distinguish between feasible and unsafe trajectories under uncertain observations.
\textbf{(4) Overall, }
these results reveal a consistent gap across capability dimensions. While models can adopt conservative fallback strategies under uncertainty, they exhibit clear limitations in observability awareness and spatial understanding. As a result, they often fail to accurately identify risk-critical regions and reliably evaluate trajectory safety. This explains the observed paradox: models \emph{know they should be cautious}, but do not fully understand \emph{what exactly is risky}.

\begin{table}[t!]
\caption{Performance of modality-specific observability across different weather conditions and sensing modalities. The best is \textbf{bold} and the second-best is \underline{underlined}.}
\label{tab:obs}
\resizebox{\columnwidth}{!}{%

\renewcommand{\arraystretch}{0.95}

\begin{tabular}{l|ccccc|ccc}
\toprule
\textbf{Model} & \textbf{Norm.} & \textbf{Rain} & \textbf{Sleet} & \textbf{Snow} & \textbf{Fog} & \textbf{CAM} & \textbf{LIDAR} & \textbf{RADAR}\\ \midrule \midrule
GPT-5 & 39.58 & 39.10 & \underline{53.63} & 48.18 & 37.84 & 48.80 &46.55  & 35.31   \\
Gemini-3.1-Pro & 56.22 & 52.13 & 49.78 & \underline{55.04} & \underline{57.36} & \underline{63.80} & 42.94 & 56.74 \\ \midrule

InternVL25-7B & 36.49 & 33.83 & 35.70 & 36.13 & 33.03 & 37.72 & 38.92  &  30.42 \\


Qwen2.5VL-7B & 41.56 & 39.10 & 42.22 & 44.12 & 34.83 & 44.14 & 37.24 & 42.54 \\
 \midrule
R1-VL-7B & 29.07 & 15.79 & 31.11 & 30.11 & 25.23 & 49.28 & 22.87 & 11.48  \\
Perception-R1-7B & 13.91 & 7.27 & 11.70 & 11.76 & 11.71 & 47.43 & 46.07 & 45.75 \\ \midrule
Dolphins & 50.71 & 48.12 & 28.44 & 22.69 & 39.64 & 32.42 & 36.36 & 51.44 \\
DriveLMM-o1 & 35.81 & 31.83 & 38.37 & 34.87 & 28.53 & 41.97 & 34.03 & 29.05 \\
Alpamayo-1.5 & 13.91 & 7.27  & 11.70 & 11.76  &  11.71 & 26.73 & 6.10 & 3.77  \\
ReasonDrive & 40.75 & 32.33 & 42.07 & 39.36 & 35.74 & 48.31 & 34.27 &  35.55 \\
RecogDrive & 33.83 & 37.59 & 34.96 & 37.25 & 35.74 & 43.90 & 24.88 & 37.00  \\ \midrule

\rowcolor[HTML]{EFEFEF}ObsDrive-SFT & \underline{65.43} & \underline{60.40} & 38.37 & 37.11 & 53.45 & 46.07  & \underline{52.17} & \underline{62.36}  \\ 
\rowcolor[HTML]{EFEFEF}ObsDrive-RL & \textbf{67.97} & \textbf{64.66} & \textbf{66.67} & \textbf{63.45} & \textbf{65.77} & \textbf{69.58} & \textbf{64.85} & \textbf{64.53}  \\
\bottomrule
\end{tabular}%
}
\end{table}

\begin{table*}[htb!]
\centering
\caption{Input-modality ablation on Spatial Reliability and Risk-aware Decision-making of ObsDriveBench.}
\label{tab:modality_ablation}
\resizebox{0.98\textwidth}{!}{%

\small
\renewcommand{\arraystretch}{0.9}
\begin{tabular}{lccc|cccc|cccc}
\toprule
Model & CAM & LIDAR & RADAR & \makecell{2D-3D \\ Ground.} 
& \makecell{Spatial \\ Relation} 
& \makecell{Interaction \\ Reasoning} & \textbf{S-Avg}  & \makecell{Risk \\ Ident.} 
& \makecell{Fallback \\ Action} 
& \makecell{Traj. \\ Eval.}  & \textbf{R-Avg} \\
\midrule
\multirow{5}{*}{Qwen2.5VL-7B}
& \ding{51} &   &   & 37.91 & 32.26 & 63.49 & 44.35 & 53.46 & 93.82 & 49.83 & 65.88 \\
& \ding{51} & \ding{51} &   & 40.96 & 33.87 & 67.22 & 47.13 & 53.22 & 94.06 & 50.17 & 65.99 \\
& \ding{51} &   & \ding{51} & 40.56 & 32.26 & 63.24 & 45.16 & 50.24 & 94.06 & 49.92 & 64.91 \\
&   & \ding{51} & \ding{51} & 38.80 & 30.42 & 65.98 & 44.83 & 44.53 & 93.58 & 46.77 & 61.80 \\
& \ding{51} & \ding{51} & \ding{51} & \textbf{44.98} & \textbf{34.91} & \textbf{67.39} &
\textbf{48.89} & \textbf{55.14} & \textbf{94.70} & \textbf{52.16} & \textbf{67.51} \\
\midrule
\multirow{2}{*}{ReCogDrive}
& \ding{51} &   &   & 37.27 & 49.92 & 67.39 & 51.35 & 51.77 & \textbf{96.39} & 65.51 & 71.29 \\
& \ding{51} & \ding{51} & \ding{51} & \textbf{39.36} & \textbf{51.44} & \textbf{69.29} &
\textbf{53.19} & \textbf{52.49} & 96.31 & \textbf{66.09} & \textbf{71.70} \\
\midrule
\multirow{2}{*}{DriveLMM-o1}
& \ding{51} &   &   & 30.44 & \textbf{49.52} & 61.74 & 47.08 & 53.38 & 93.58 & \textbf{49.59} & 65.69 \\
& \ding{51} & \ding{51} & \ding{51} & \textbf{30.76} & \textbf{49.52} & \textbf{63.40} &
\textbf{47.73} & \textbf{53.94} & \textbf{93.66} & 49.25 & \textbf{65.80} \\
\bottomrule
\end{tabular}}
\end{table*}

\begin{table*}[htb!]
\centering
\caption{Cross-dataset evaluation on DriveBench under different weather. 
All results are accuracy (\%); bold indicates the better result.}
\label{tab:drivebench}
\small
\resizebox{0.98\textwidth}{!}{%

\renewcommand{\arraystretch}{0.9}
\begin{tabular}{lcc cc cc cc}
\toprule
\multirow{2}{*}{Weather}
& \multicolumn{2}{c}{Perception}
& \multicolumn{2}{c}{Prediction}
& \multicolumn{2}{c}{Planning}
& \multicolumn{2}{c}{Behavior} \\
\cmidrule(lr){2-3}
\cmidrule(lr){4-5}
\cmidrule(lr){6-7}
\cmidrule(lr){8-9}
& Qwen2.5VL-7B & ObsDrive
& Qwen2.5VL-7B & ObsDrive
& Qwen2.5VL-7B & ObsDrive
& Qwen2.5VL-7B & ObsDrive \\
\midrule
Clean
& 42.8 & \textbf{42.9}
& 35.4 & \textbf{35.8}
& 46.2 & \textbf{47.5}
& 60.5 & \textbf{61.1} \\
Dark
& 36.2 & \textbf{37.6}
& 33.4 & \textbf{34.9}
& 40.4 & \textbf{41.5}
& 60.5 & \textbf{61.5} \\
Fog
& 41.5 & \textbf{43.0}
& 34.5 & \textbf{36.9}
& 45.1 & \textbf{45.4}
& 56.8 & \textbf{57.9} \\
Rain
& 40.1 & \textbf{41.5}
& 36.1 & \textbf{37.2}
& 45.9 & \textbf{46.2}
& 58.9 & \textbf{59.7} \\
Snow
& 41.9 & \textbf{42.8}
& 37.2 & \textbf{38.1}
& 40.6 & \textbf{41.6}
& 59.4 & \textbf{61.0} \\
\bottomrule
\end{tabular}
}
\end{table*}

\begin{table}[htb!]
\centering
\caption{Cross-dataset results on DriveLMM-o1. Each cell reports
accuracy/reasoning score following its official protocol; bold denotes the better result.}
\label{tab:drivelmm_external}
\small
\setlength{\tabcolsep}{4pt}

\resizebox{0.98\columnwidth}{!}{%

\renewcommand{\arraystretch}{0.9}
\begin{tabular}{@{}llcc@{}}
\toprule
Model & Input & Night & Rain \\
\midrule
\multirow{2}{*}{Qwen2.5VL-7B}
& CAM       & 64.19 / 74.13 & 55.92 / 71.58 \\
& CAM+LiDAR & \textbf{73.02 / 77.03} & \textbf{62.88 / 73.32} \\
\midrule
\multirow{2}{*}{ObsDrive}
& CAM       & 64.52 / 74.23 & 54.29 / 71.34 \\
& CAM+LIDAR & \textbf{73.95 / 77.39} & \textbf{64.10 / 74.04} \\
\midrule
\multirow{2}{*}{DriveLMM-o1}
& CAM       &  \textbf{71.63 / 75.88} & \textbf{61.72} / 74.01 \\
& CAM+LIDAR & 71.44 / 75.76 &  61.48 / \textbf{74.16} \\
\bottomrule
\end{tabular}
}
\end{table}

\subsubsection{Observability Awareness.} 

Table~\ref{tab:obs} presents detailed results of modality-specific observability prediction across different weather conditions and sensing modalities.
\textbf{(1) Impact of adverse weather.}
We first observe a clear performance gap between normal and adverse weather conditions. Most models achieve their best performance under normal weather, while performance degrades under rain, snow, sleet, and fog. This confirms that adverse weather significantly impairs models' ability to understand environmental observability. This degradation indicates that current MLLMs lack robustness when observations become unreliable. In contrast, our model (ObsDrive-RL) maintains relatively stable performance across all weather conditions, suggesting improved robustness to observability degradation.
\textbf{(2) Modality imbalance.}
Across all methods, performance on camera-based observability is consistently higher than that on lidar and radar, highlighting a strong modality bias in current models. In particular, radar shows the lowest performance for most models, reflecting the difficulty of interpreting sparse and coarse representations. This suggests that existing models are not well equipped to understand about modality-dependent observability, especially when non-visual modalities become critical under adverse weather.
\textbf{(3) Effect of training strategy.}
Comparing Ours-SFT and Ours-RL, we observe that reinforcement learning under adverse-weather scenarios leads to substantial improvements across all weather conditions and modalities. Notably, Ours-RL not only improves performance under adverse weather but also enhances generalization to normal conditions. This suggests that explicitly training under degraded and uncertain observations helps the model better calibrate its observability understanding.
\textbf{(4) Implications.}
Overall, these results demonstrate that modality-specific observability is a challenging yet essential capability that is not well captured by existing benchmarks. The observed performance gaps across weather conditions and modalities highlight the importance of explicitly modeling observability and training models to reason under degraded and cross-modal observations.

\subsection{Input-modality Ablation}

Table~\ref{tab:modality_ablation} examines whether adding LiDAR/radar BEV inputs causes feature-distribution mismatch in vision-centric VLMs. Compared with camera-only input, using all three modalities improves S-AVG by 4.54, 1.84, and 0.65 points for Qwen2.5VL-7B, ReCogDrive, and DriveLMM-o1, respectively, while R-AVG improves by 1.63, 0.41, and 0.11 points. Thus, BEV inputs do not cause systematic degradation in overall performance. Nevertheless, improvements are task dependent, and several individual trajectory or fallback scores change marginally or slightly decrease.

\subsection{Cross-dataset Evaluation}
\label{sec:drivebench}

No existing public benchmark provides the same real adverse-weather camera-LiDAR-radar observability annotations as ObsDriveBench, making exact task-level cross-dataset evaluation unavailable. We therefore conduct two
complementary evaluations on independent nuScenes. First, on DriveBench~\cite{xie2025vlms} with camera-only synthetic corruptions, ObsDrive outperforms Qwen2.5-VL-7B in all 20 task-condition settings (Table~\ref{tab:drivebench}). This demonstrates improvements transfer across scenes and corruption patterns.
We further evaluate on real nighttime and rainy subsets of the DriveLMM-o1 test set~\cite{ishaq2025drivelmm}, which differs from K-Radar in scene and sensor distributions. With camera and LiDAR, ObsDrive outperforms Qwen2.5VL-7B by 0.93/0.36 accuracy/reasoning scores at night and 1.22/0.72 in rain (Table~\ref{tab:drivelmm_external}). DriveLMM-o1 is fine-tuned on its own training split with image-only inputs, and adding LiDAR BEV does not consistently improve its results. Despite no training on DriveLMM-o1 dataset, multimodal ObsDrive surpasses it by 2.51/1.63 points at night and 2.62 accuracy in rain, while achieving comparable rain reasoning (74.04 vs.\ 74.16). These results support multimodal transfer beyond K-Radar, while camera-only robustness under rain remains a limitation.

\section{Conclusion}
\label{sec_conclusion}

In this work, we introduce \textbf{ObsDriveBench}, a real-world multi-modal benchmark for evaluating autonomous driving under adverse weather. ObsDriveBench provides a structured evaluation along three complementary dimensions: observability awareness, spatial reliability, and risk-aware decision-making, enabling fine-grained diagnosis of model behavior under degraded observations.
Our experiments reveal systematic limitations of current models, including inaccurate observability estimation and inconsistent spatial understanding. While models often adopt conservative fallback strategies, they frequently fail to precisely identify risk-critical regions and evaluate trajectory safety, exposing a gap between uncertainty awareness and effective decision-making.
We further show that explicitly modeling observability and uncertainty improves robustness across all capability dimensions. ObsDriveBench provides a valuable testbed for future research on reliable multimodal understanding and safe decision-making.

{
    \small
    \bibliographystyle{ieeenat_fullname}
    \bibliography{main}
}

\end{document}